\documentclass{article}
\usepackage{iclr2027_conference,times}
\usepackage{amsmath,amsfonts,bm}

\def\eqref#1{equation~\ref{#1}}
\def\1{\bm{1}}

\DeclareMathAlphabet{\mathsfit}{\encodingdefault}{\sfdefault}{m}{sl}
\SetMathAlphabet{\mathsfit}{bold}{\encodingdefault}{\sfdefault}{bx}{n}

\usepackage{amsmath,amssymb,mathtools}
\newtheorem{proposition}{Proposition}
\newtheorem{corollary}{Corollary}
\newtheorem{lemma}{Lemma}
\usepackage{booktabs,array,tabularx,adjustbox}
\usepackage{graphicx,float,microtype}
\usepackage{algorithm}
\usepackage{algpseudocode}

\floatstyle{ruled}
\restylefloat{algorithm}

\algrenewcommand{\algorithmicrequire}{\textbf{Input:}}
\algrenewcommand{\algorithmicensure}{\textbf{Output:}}
\algrenewcommand{\algorithmicindent}{1.2em}
\usepackage{xcolor}
\usepackage{colortbl}
\usepackage{hyperref}
\usepackage{wrapfig}
\usepackage{needspace}
\usepackage{capt-of}

\newsavebox{\upwrapbox}

\usepackage{url}
\hypersetup{hidelinks}
\definecolor{upteal}{HTML}{087F75}
\definecolor{upgray}{HTML}{5F6B76}
\definecolor{uprow}{HTML}{E8F6F3}
\title{UP-MOPD: Update Projection in Multi-Teacher On-Policy Distillation}
\author{%
{\bfseries Taojie Zhu\textsuperscript{1,2,*}\quad
Jing Jin\textsuperscript{1,*}}\\[3pt]
{\normalfont Yuan Xia\textsuperscript{2,$\dagger$}\quad
Chenyang Ding\textsuperscript{1}\quad
Qunshan He\textsuperscript{2,3}\quad
Wanke Xia\textsuperscript{1}}\\[3pt]
{\normalfont Tao Sun\textsuperscript{2}\quad
Yan Chen\textsuperscript{1}\quad
Jian Wang\textsuperscript{2}\quad
Jinjie Gu\textsuperscript{2}\quad
Tao Feng\textsuperscript{1,$\dagger$}}\\[7pt]
{\normalfont\small\mbox{\textsuperscript{1}Tsinghua University\quad
\textsuperscript{2}Ant Group\quad
\textsuperscript{3}Zhejiang University}}\\[5pt]
{\normalfont\small\textsuperscript{$\dagger$}\href{mailto:zhongjing.xy@antgroup.com}{\texttt{zhongjing.xy@antgroup.com}}\quad
\href{mailto:fengtao.hi@gmail.com}{\texttt{fengtao.hi@gmail.com}}}%
}
\date{}
\iclrfinalcopy
\hypersetup{
  pdftitle={UP-MOPD: Update Projection in Multi-Teacher On-Policy Distillation},
  pdfauthor={Taojie Zhu, Jing Jin, Yuan Xia, Chenyang Ding, Qunshan He, Wanke Xia, Tao Sun, Yan Chen, Jian Wang, Jinjie Gu, Tao Feng},
  pdfsubject={Multi-teacher on-policy distillation},
  pdfkeywords={on-policy distillation, multi-teacher learning, update projection, capability integration}
}

\begin{document}
\raggedbottom
\maketitle
% Remove the final-copy publication banner: this version is a public preprint.
\fancyhf{}
\fancyfoot[C]{\thepage}
\renewcommand{\headrulewidth}{0pt}
\renewcommand{\footrulewidth}{0pt}
\pagestyle{fancy}
\thispagestyle{fancy}
\begingroup
\renewcommand{\thefootnote}{\fnsymbol{footnote}}
\footnotetext[1]{Equal contribution. This work was completed during internship at Ant Group.}
\endgroup

\begin{abstract}
On-policy distillation from multiple teachers combines expertise from different domains in a single student, but conflicting gradients can hinder this integration.
Gradient corrections directly constrain parameter updates under plain SGD.
With optimizers such as AdamW, however, momentum, adaptive scaling, and weight decay can turn a corrected gradient into an update that increases a domain loss to first order.
To address this gap, we propose \textbf{U}pdate \textbf{P}rojection for \textbf{M}ulti-Teacher \textbf{O}n-\textbf{P}olicy \textbf{D}istillation (\textbf{UP-MOPD}).
UP-MOPD lets the original mixed gradient update the optimizer state and generate a candidate displacement, then projects only violating candidates before they are committed to the parameters.
The projection gives the unique feasible update closest to the candidate in Euclidean distance.
In experiments combining medical and general domains, UP-MOPD improves IFEval-loose accuracy late in training by 2.96 points over vanilla M-OPD.
It achieves an average score of 60.03 across eight metrics, compared with 59.00 for gradient projection and 59.15 for update rejection.
On a public benchmark covering mathematics, code, and instruction following, it achieves the best average across six tasks (32.67), leads on LiveCodeBench v5, and ties for the best IFEval result.
These results support projecting optimizer updates to reduce interference between domains. Our code is available at: \url{https://anonymous.4open.science/r/UP-MOPD/}
\end{abstract}

\section{Introduction}

Large-model post-training commonly produces experts specialized for different capabilities or domains. Two forms of consolidation are particularly useful in practice: combining foundational capabilities such as mathematics, code, and instruction following, and merging such general capabilities with vertical expertise in medicine, finance, or law. Multi-teacher on-policy distillation (M-OPD) provides a direct route to both forms of integration: the student generates its own trajectories, and the corresponding teachers provide dense token-level supervision. MiMo-V2-Flash and subsequent MOPD work adopt the pattern of training domain teachers separately and distilling them into a shared student \citep{ref31,ref32}; GLM-5 uses a related on-policy cross-stage distillation procedure to mitigate capability forgetting during sequential post-training \citep{ref33}.

Existing M-OPD studies primarily improve the construction and allocation of teacher signals. Open-MOPD corrects effective optimization-budget imbalance arising from sequence length, teacher--student gaps, and reward staleness \citep{ref3}; related work organizes multi-task teachers through privileged context \citep{ref23} or studies behavioral drift caused by insufficient supervision of decision-critical tokens \citep{ref24}. Yet after routing, supervision, and domain budgets have been fixed, all domain losses still share one parameter vector and one optimizer. M-OPD first forms the mixed gradient $g_0=\sum_d q_dg_d$; AdamW then combines it with historical moments, coordinate-wise scaling, and weight decay to produce a candidate parameter displacement $\Delta\theta_0$, namely the update proposed by the optimizer before parameter commit. This exposes a stage not covered by gradient-level interventions: even when $-g_0$ is first-order non-increasing for every domain in gradient space, $\Delta\theta_0$ may still increase the current loss of one domain. This raises a direct question: how can a harmful optimizer-produced displacement be corrected while minimally changing the intended update?

\begin{figure}[t]
\centering
\includegraphics[width=\linewidth]{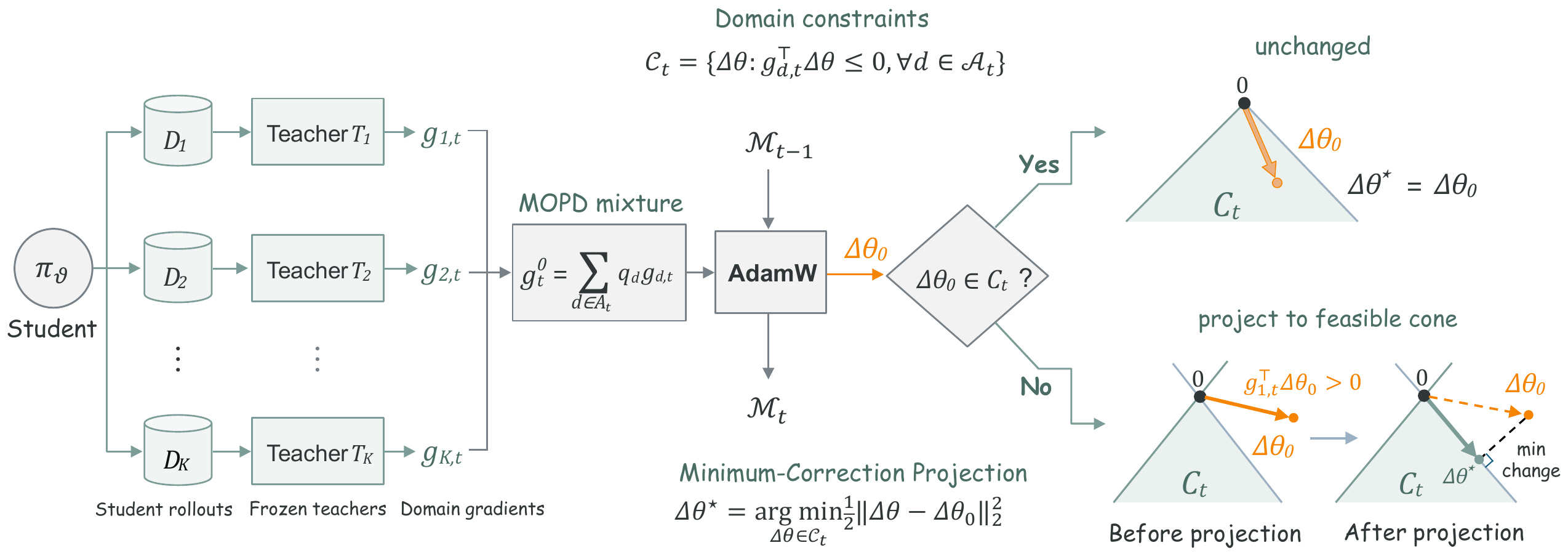}
\caption{Overview of UP-MOPD. AdamW forms a candidate parameter displacement
from the mixed distillation gradient. UP-MOPD evaluates its first-order effect
on each active domain and minimally corrects a violating candidate before
commit, while retaining the current-step optimizer state produced by the
original mixture.}
\label{fig:overview}
\end{figure}

Multi-task learning methods reduce interference by modifying task gradients or adjusting their weights before the optimizer step \citep{ref25,ref26,ref34,ref27,ref28}.
PCGrad, for example, projects conflicting task gradients before combining them.
We formulate \textbf{G}radient \textbf{P}rojection for \textbf{M}ulti-Teacher \textbf{O}n-\textbf{P}olicy \textbf{D}istillation (\textbf{GP-MOPD}), which jointly projects the mixed gradient before the optimizer step.
Under plain SGD, correcting the gradient directly controls the parameter update.
With AdamW, however, momentum, adaptive scaling, and weight decay can turn a corrected gradient into an update that increases a domain loss to first order \citep{ref4,ref5}.
To address this mismatch, we propose \textbf{U}pdate \textbf{P}rojection for \textbf{M}ulti-Teacher \textbf{O}n-\textbf{P}olicy \textbf{D}istillation (\textbf{UP-MOPD}), which directly corrects the parameter update produced by AdamW before applying it to the model.

For domain $d$, the current-batch first-order loss change induced by a candidate displacement $\Delta\theta$ is
\[
h_d(\Delta\theta)=g_d^\top\Delta\theta.
\]
Under a loss-minimization convention, $h_d>0$ denotes first-order harm. A constructive counterexample shows that positive diagonal preconditioning can change gradient-space feasibility (Proposition~\ref{prop:precondition}; Appendix~\ref{app:proof_precondition}), while actual training trajectories show that $g_d^\top(-g_0)$ and $g_d^\top\Delta\theta_0$ do not always have the same sign (Section~\ref{sec:exp_mechanism}). Consequently, a gradient-only criterion may either miss an optimizer-induced harmful update or intervene unnecessarily on an update that is safe after optimization.

UP-MOPD minimally corrects AdamW's candidate update to satisfy all active domains' first-order constraints while retaining the optimizer state computed from the original mixed gradient.

Motivated by the substantial gap between medical and general domains, we investigate interference when consolidating their capabilities into a shared student. We analyze the intervention through controlled comparisons and evaluate the method on a public benchmark covering mathematics, code, and instruction following.

Our contributions are threefold:
\begin{enumerate}

    \item We show that gradient-space feasibility does not guarantee a feasible parameter update in M-OPD. We prove this non-invariance constructively and verify gradient displacement disagreement in AdamW trajectories.

    \item We formulate GP-MOPD in gradient space and propose UP-MOPD at the optimizer output. UP-MOPD gives the unique minimum correction satisfying all domain constraints while retaining the current-step optimizer state, with a dual dimension determined only by the active domains.

    \item Across medical--general and public math--code--instruction consolidation, UP-MOPD consistently outperforms GP-MOPD in aggregate, supporting commit-side projection across domain compositions, model families, and M-OPD implementations.

\end{enumerate}

\section{Method}

Inspired by PCGrad~\citep{ref26}, GP-MOPD projects the mixed distillation
gradient before optimization. Because AdamW subsequently incorporates
optimizer history, coordinate-wise scaling, and weight decay, its resulting
displacement can violate constraints satisfied in gradient space. UP-MOPD
instead lets AdamW form the candidate from the original mixture and applies
the minimum correction required by all active domains
(Figure~\ref{fig:overview}).

\subsection{M-OPD Objective and Optimizer Candidate}

At training step $t$, let $\mathcal A_t$ denote the active domains,
$D_t=|\mathcal A_t|$, and $\mathcal I_{d,t}=\{i:d_i=d\}$ the samples assigned
to domain $d$. For each input $x_i$, the frozen rollout-policy snapshot
$\pi_{\vartheta_t}$ generates
\[
y_i=(y_{i,1},\ldots,y_{i,T_i})
\sim\pi_{\vartheta_t}(\cdot\mid x_i).
\]
At position $k$, teacher $\pi_{\phi_d}$ supervises the student-visited state
$s_{i,k}=(x_i,y_{i,<k})$. In the sampled-token form, we define
\begin{equation}
a_{i,k}^{(d)}
=\operatorname{sg}\!\left[
\log\pi_{\phi_d}(y_{i,k}\mid s_{i,k})-
\log\pi_{\vartheta_t}(y_{i,k}\mid s_{i,k})\right],
\qquad
\rho_{i,k}(\theta)=
\frac{\pi_\theta(y_{i,k}\mid s_{i,k})}
     {\pi_{\vartheta_t}(y_{i,k}\mid s_{i,k})}.
\label{eq:mopd_advantage}
\end{equation}
Let $m_{i,k}$ be the valid-token mask, $M_i=\sum_km_{i,k}>0$, and
$Z_d=\sum_{i\in\mathcal I_{d,t}}\omega_iM_i$. The domain loss is
\begin{equation}
\begin{aligned}
L_d(\theta)=-\frac{1}{Z_d}
\sum_{i\in\mathcal I_{d,t}}\omega_i
\sum_{k=1}^{T_i}m_{i,k}
\min\!\Big[
&\rho_{i,k}(\theta)a_{i,k}^{(d)},
&\operatorname{clip}(\rho_{i,k}(\theta),1-\xi,1+\xi)a_{i,k}^{(d)}
\Big],
\end{aligned}
\label{eq:mopd_domain_loss}
\end{equation}
where $\xi$ is the policy-ratio clipping threshold; $\omega_i=1/M_i$ and
$\omega_i=1$ give response- and token-mean normalization, respectively. The
public implementation uses the dense top-$k$ counterpart of
Equation~(\ref{eq:mopd_domain_loss}); UP-MOPD only requires differentiable,
domain-decomposable losses (Appendix~\ref{app:domain_replay}).

For coefficients $q_d>0$, treated as constants during the current step and
satisfying $\sum_{d\in\mathcal A_t}q_d=1$, define
\begin{equation}
\begin{aligned}
g_d&=\nabla_\theta L_d(\theta_t),\qquad
\widetilde g_d=q_dg_d,\\
g_0&=\sum_{d\in\mathcal A_t}\widetilde g_d
=\nabla_\theta\!\left[
\sum_{d\in\mathcal A_t}q_dL_d(\theta)
\right]_{\theta=\theta_t}.
\end{aligned}
\label{eq:up_mixture}
\end{equation}
The mixed pass produces $g_0$; replaying the same responses at $\theta_t$
produces the constraint gradients $g_d$ with matched masks, normalization,
and stochastic state. Any global clipping scale is computed from the mixture
and applied uniformly (Appendix~\ref{app:domain_replay}).

Standard M-OPD passes $g_0$ to the shared optimizer:
\begin{equation}
\begin{aligned}
(\theta_{t+1}^{0},\mathcal M_t)
&=\operatorname{Opt}_t(\theta_t,\mathcal M_{t-1};g_0),\\
\Delta\theta_0&=\theta_{t+1}^{0}-\theta_t.
\end{aligned}
\label{eq:up_candidate}
\end{equation}
The candidate displacement $\Delta\theta_0$ is the complete update proposed
before parameter commit and includes AdamW moments, adaptive scaling, weight
decay, and any shared scalar clipping.

Ignoring arithmetic rounding, the standard AdamW candidate has the form
\begin{equation}
\Delta\theta_0
=-\eta_t\frac{\widehat m_t}{\sqrt{\widehat v_t}+\varepsilon}
-\eta_t\lambda_{\mathrm{wd}}\theta_t,
\qquad
(\widehat m_t,\widehat v_t)=F_t(g_0^{\mathrm{opt}};\mathcal M_{t-1}),
\label{eq:adamw_candidate}
\end{equation}
where $g_0^{\mathrm{opt}}$ is the gradient presented to AdamW after any shared
scalar clipping and equals $g_0$ when clipping is disabled. Division is
coordinate-wise, and $F_t$ denotes moment accumulation and bias correction.
Thus $\Delta\theta_0$ depends on the optimizer history and the full
decoupled-decay term rather than being a scalar multiple of $g_0$. In the
implementation, we obtain the candidate by differencing the FP32 master
parameters before and after the optimizer call, which also covers fused update
kernels and the shared clipping path.

\subsection{GP-MOPD and the Optimizer-Output Mismatch}
\label{sec:gp_mopd}

PCGrad performs sequential pairwise projections before aggregation. GP-MOPD
instead corrects the mixed gradient in one joint projection by requiring
nonnegative alignment with every domain gradient:
\begin{equation}
g_{\mathrm{GP}}^\star
=\operatorname*{arg\,min}_{g\in\mathbb R^P}
\frac12\|g-g_0\|_2^2
\quad\text{s.t.}\quad
g_d^\top g\ge0,\quad d\in\mathcal A_t.
\label{eq:gp_projection}
\end{equation}
The projected gradient is then passed to the original optimizer,
\[
(\theta_{t+1}^{\mathrm{GP}},\mathcal M_t^{\mathrm{GP}})
=\operatorname{Opt}_t(\theta_t,\mathcal M_{t-1};g_{\mathrm{GP}}^\star).
\]
Under scalar-step SGD the two interventions are equivalent, as formalized in
Section~\ref{sec:up_mopd}; Equation~(\ref{eq:adamw_candidate}) shows why this
need not hold for AdamW.

\begin{proposition}[Diagonal preconditioning can destroy feasibility]
\label{prop:precondition}
There exist domain gradients $\{g_d\}$, positive mixture weights $q_d$ with
$\sum_dq_d=1$, and a positive diagonal matrix $D\succ0$ such that, for
$g_0=\sum_dq_dg_d$, the unpreconditioned displacement is strictly feasible,
$g_d^\top(-g_0)<0$ for every $d$, while the preconditioned displacement
violates at least one constraint, $g_j^\top(-Dg_0)>0$ for some $j$.
\end{proposition}
For equal mixture weights, a concrete construction is
\begin{equation}
g_1=(2,-1)^\top,\qquad g_2=(0,3)^\top,\qquad g_0=(1,1)^\top.
\label{eq:precondition_example}
\end{equation}
Here $g_1^\top(-g_0)=-1$ and $g_2^\top(-g_0)=-3$, whereas
$D=\operatorname{diag}(0.1,1)$ gives $g_1^\top(-Dg_0)=0.8>0$.
Appendix~\ref{app:proof_precondition} gives the full verification. This
existence result isolates positive diagonal scaling rather than claiming that
every AdamW step violates a domain constraint. Whether a particular step is
feasible must be decided from its candidate displacement $\Delta\theta_0$.

\subsection{UP-MOPD: Minimum Correction of the Candidate Update}
\label{sec:up_mopd}

The domain gradients estimate how AdamW's candidate changes the current-batch
losses. If $L_d$ has a locally Lipschitz gradient on the segment between
$\theta_t$ and $\theta_t+\Delta\theta$, then
\begin{equation}
\begin{aligned}
L_d(\theta_t+\Delta\theta)-L_d(\theta_t)
&=g_d^\top\Delta\theta+O(\|\Delta\theta\|_2^2),\\
h_d&=g_d^\top\Delta\theta_0.
\end{aligned}
\label{eq:up_harm}
\end{equation}
A value $h_d>0$ is a first-order violation: the candidate is predicted to
increase domain $d$'s loss. Because correcting one domain can change the
effects on others, UP-MOPD enforces all constraints jointly. Writing
$G=[g_d]_{d\in\mathcal A_t}\in\mathbb R^{P\times D_t}$, we solve
\begin{equation}
\Delta\theta^\star
=\operatorname*{arg\,min}_{\Delta\theta\in\mathbb R^P}
\frac12\|\Delta\theta-\Delta\theta_0\|_2^2
\quad\text{s.t.}\quad
G^\top\Delta\theta\le\mathbf0.
\label{eq:up_projection}
\end{equation}
The inequalities define half-spaces whose intersection contains zero.
Joint projection avoids ordering effects: it leaves feasible candidates
unchanged and otherwise returns the nearest feasible displacement.

\begin{lemma}[Equivalence under scalar-step SGD]
\label{lem:sgd_equivalence}
If the optimizer applies $\Delta\theta=-\eta_t g$ with $\eta_t>0$, then the
gradient projection in Eq.~(\ref{eq:gp_projection}) and the displacement
projection in Eq.~(\ref{eq:up_projection}) are equivalent:
$\Delta\theta^\star=-\eta_tg_{\mathrm{GP}}^\star$.
\end{lemma}
Indeed, substituting $\Delta\theta=-\eta_tg$ maps
$g_d^\top\Delta\theta\le0$ exactly to $g_d^\top g\ge0$ and scales the squared
distance objective by the positive constant $\eta_t^2$. The lemma isolates the
source of the distinction between GP-MOPD and UP-MOPD: it arises from the
optimizer transformation, not from different domain constraints.

\begin{proposition}[Unique minimum correction]
\label{prop:projection}
Let $\mathcal C_t=\{\Delta\theta:G^\top\Delta\theta\le\mathbf0\}$. The set
$\mathcal C_t$ is a nonempty closed convex cone, and
Eq.~(\ref{eq:up_projection}) has a unique solution. It leaves a feasible
candidate unchanged and otherwise returns the nearest feasible displacement.
For a fixed candidate, the feasible cone and its exact projection are invariant
to independent positive rescaling of the constraint normals. The projection
also satisfies
\begin{equation}
\|\Delta\theta_0\|_2^2
=\|\Delta\theta^\star\|_2^2
+\|\Delta\theta_0-\Delta\theta^\star\|_2^2.
\label{eq:up_pythagorean}
\end{equation}
\end{proposition}
The feasible set is closed and convex, while the squared-distance objective is
coercive and strongly convex, which gives existence and uniqueness.
Equation~(\ref{eq:up_pythagorean}) orthogonally decomposes the candidate into
the feasible displacement that is retained and the normal correction that is
removed; in particular,
$\|\Delta\theta^\star\|_2\le\|\Delta\theta_0\|_2$. The full proof is given in
Appendix~\ref{app:proof_projection}.

\begin{corollary}[First-order consistency]
\label{cor:mixed_first_order}
Every feasible projected displacement satisfies
$g_0^\top\Delta\theta^\star=
\sum_{d\in\mathcal A_t}q_dg_d^\top\Delta\theta^\star\le0$.
If $L_d$ is $\beta_d$-smooth on the segment between $\theta_t$ and
$\theta_t+\Delta\theta^\star$, then
\begin{equation}
L_d(\theta_t+\Delta\theta^\star)-L_d(\theta_t)
\le\frac{\beta_d}{2}\|\Delta\theta^\star\|_2^2.
\label{eq:finite_step_bound}
\end{equation}
The constraints therefore certify current-batch first-order non-increase, not
domain-wise monotonicity for a finite update.
\end{corollary}
The proof is given in Appendix~\ref{app:proof_projection}.

UP-MOPD directly commits $\Delta\theta^\star$ while retaining the current-step
state $\mathcal M_t$ computed from the original mixture. Subsequent states
evolve along the modified parameter trajectory and need not match unmodified
AdamW.

\subsection{Domain-Dimensional Dual and Training Integration}

Introducing multipliers $\lambda\in\mathbb R_+^{D_t}$ gives
$\mathcal L(\Delta\theta,\lambda)=
\tfrac12\|\Delta\theta-\Delta\theta_0\|_2^2+
\lambda^\top G^\top\Delta\theta$. Stationarity yields
\begin{equation}
\Delta\theta^\star=\Delta\theta_0-G\lambda^\star.
\label{eq:up_correction}
\end{equation}
The correction therefore depends on one nonnegative coefficient per active
domain. Defining $K=G^\top G$ and $h=G^\top\Delta\theta_0$, substitution gives
the low-dimensional dual
\begin{equation}
\lambda^\star
=\operatorname*{arg\,min}_{\lambda\ge\mathbf0}
\frac12\lambda^\top K\lambda-\lambda^\top h.
\label{eq:up_dual}
\end{equation}
Optimal pairs satisfy
\begin{equation}
G^\top\Delta\theta^\star\le\mathbf0,\qquad
\lambda^\star\ge\mathbf0,\qquad
\lambda^\star\odot(G^\top\Delta\theta^\star)=\mathbf0.
\label{eq:up_kkt}
\end{equation}
Here $h$ contains the candidate's linearized domain effects and $K$ the
geometry of the constraint normals. Positive multipliers correspond to binding
constraints. Even if singular $K$ makes $\lambda^\star$ nonunique,
$G\lambda^\star$ and the primal solution remain unique. The correction therefore
lies in the active-gradient span; Appendix~\ref{app:dual_solver} gives the
active-set solver and distributed Gram construction.

The training path follows directly from these quantities. A single-domain step
uses standard M-OPD. On a multi-domain step, a feasible AdamW candidate is
committed unchanged; only a violating candidate invokes the $D_t$-dimensional
dual solve. Because the projection is computed before finite-precision
materialization, the realized FP32 displacement is checked again against the
same constraints. A bounded repair is attempted if rounding reintroduces a
violation, and the zero displacement provides an always-feasible safety
fallback. Algorithm~\ref{alg:up_mopd} and the materialization procedure are
given in Appendices~\ref{app:implementation} and
\ref{app:finite_precision}, respectively.

Beyond the $D_t$ domain-gradient replay passes, streaming construction of the
Gram statistics requires $O(D_t^2P)$ inner-product work and communicates only
$O(D_t^2)$ scalars. Full storage, solver, and verification costs are detailed
in Appendices~\ref{app:dual_solver}--\ref{app:complexity}.

\section{Experiments}

We evaluate GP-MOPD and UP-MOPD in medical and general capability integration and on a public benchmark covering mathematics, code, and instruction following. We then compare intervention strategies and analyze training trajectories in the medical setting to examine the design of update projection.

\subsection{Experimental Setup}

We train for one epoch in both settings; training parameters are given in Appendix~\ref{app:exp_training}.

\paragraph{Medical and General Capability Integration.}
We use Qwen3-4B-Instruct-2507~\citep{ref6} as the student initialization and general teacher, alongside a medical teacher trained following the recipe of \citet{zhu2026conrubmed}. The medical teacher supervises 5,166 examples from RaR-Medicine~\citep{gunjal2025rubrics}, and the general teacher supervises DAPO-Math-17k~\citep{ref7}. The two datasets are combined and randomly shuffled. We compare GP-MOPD and UP-MOPD with M-OPD, with additional comparisons of gradient clipping, domain reweighting, and update rejection in Section~\ref{sec:exp_mechanism}.

We evaluate medical capabilities on HealthBench-Hard~\citep{ref8}, MedMCQA~\citep{ref11}, PubMedQA~\citep{ref12}, and MMLU-Med~\citep{ref13}, and general capabilities on GPQA~\citep{ref9}, IFEval~\citep{ref10}, and AIME24/25. Results average the final three checkpoints, saved at 200-step intervals, with prompt-level loose accuracy for IFEval and avg@5 for GPQA and AIME.

\begin{wrapfigure}[22]{l}{0.67\linewidth}
  \centering
  \vspace{-15pt} %
  \includegraphics[width=\linewidth]{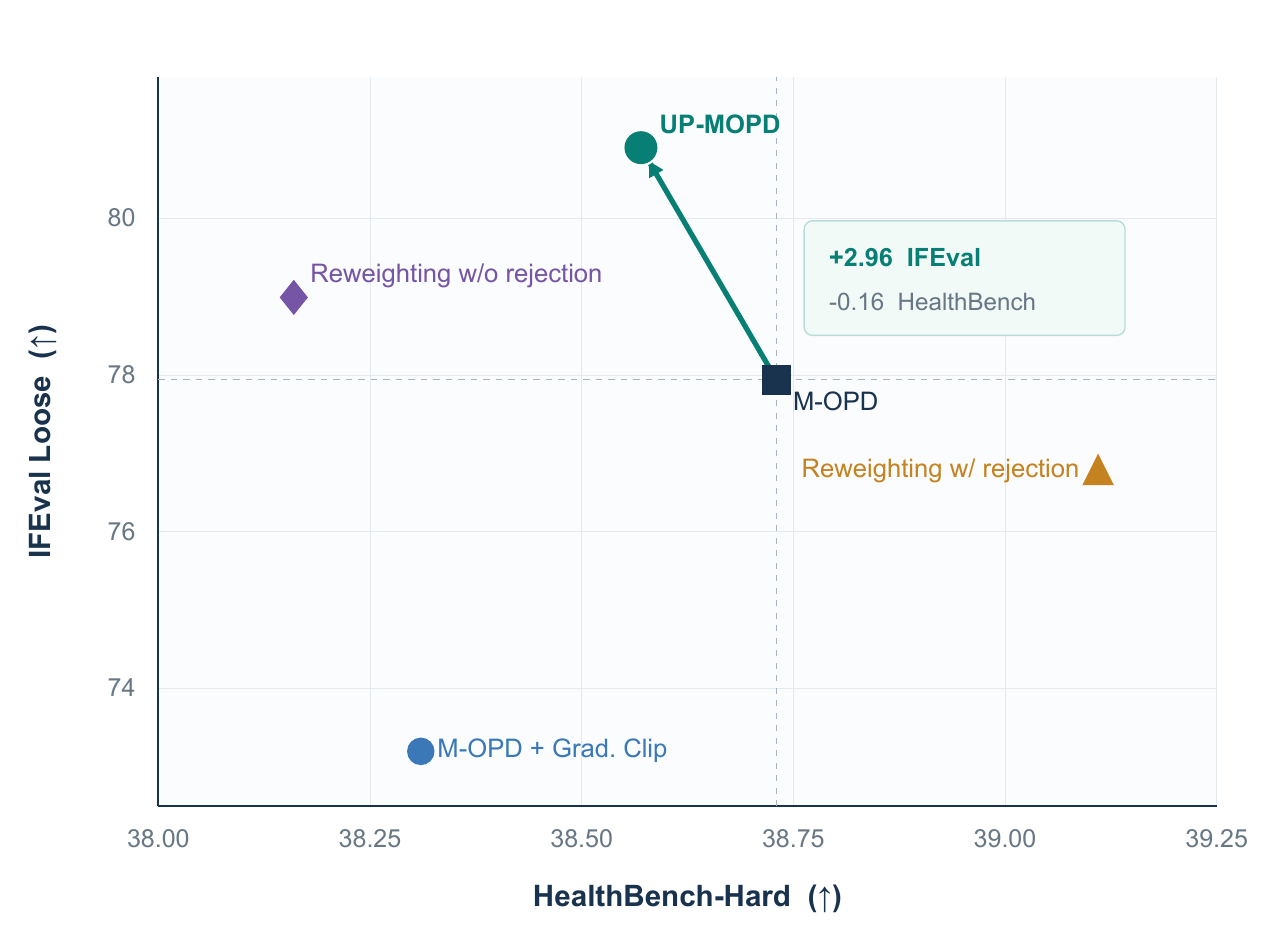}
  \caption{Medical performance and instruction following. Student points average the last three saved checkpoints of one training epoch. The arrow from M-OPD to UP-MOPD shows improved IFEval-loose accuracy at a similar HealthBench-Hard score.}
  \label{fig:capability}
  \vspace{-20pt} %
\end{wrapfigure}

\paragraph{Integration of Mathematics, Code, and Instruction Following.}
We adopt the public data and checkpoints from Open-MOPD~\citep{ref3}, using the SmolLM3-3B MixSFT student and three RL teachers. Baselines include data-based consolidation (RFT), mixed-domain RL (MixRL), parameter merging (ParamMerge-Avg and ParamMerge-TA), and distillation methods (Naive M-OPD, Open-MOPD, and RouteOPD). Teacher and initial-student results provide additional references. Quoted results are marked with $^*$ in Table~\ref{tab:public}.

We evaluate the final checkpoint on AIME24/25, LiveCodeBench v5/v6~\citep{ref14}, IFEval, and IFBench under the Open-MOPD protocol: AIME24/25 use mean@64, LiveCodeBench v5/v6 use mean@10, and IFEval and IFBench use mean@1.

\subsection{Main Results}
\label{sec:exp_main}

\paragraph{Medical and General Capability Integration.}
Table~\ref{tab:1} shows that UP-MOPD achieves the highest student average across the eight metrics at 60.03. Relative to M-OPD, UP-MOPD improves IFEval-loose accuracy by 2.96 percentage points and PubMedQA accuracy by 2.66 points. Its HealthBench-Hard score reaches 38.57, close to the medical teacher's 39.06. Figure~\ref{fig:capability} illustrates the improvement in instruction following while maintaining a similar level of HealthBench-Hard performance.

\begin{table}[t]
\centering
\caption{Medical and general capability integration. Student results are averaged over the final three checkpoints. Mean averages the eight metrics. Bold indicates the best student result in each column.}
\label{tab:1}
\renewcommand{\arraystretch}{1.08}
\scriptsize
\setlength{\tabcolsep}{3pt}
\begin{tabular*}{\linewidth}{@{\extracolsep{\fill}}lccccccccc@{}}
\toprule
& \multicolumn{4}{c}{\textbf{Medical}}
& \multicolumn{2}{c}{\textbf{General}}
& \multicolumn{2}{c}{\textbf{Math}} & \\
\cmidrule(lr){2-5}\cmidrule(lr){6-7}\cmidrule(lr){8-9}
\textbf{Method}
& \shortstack{\textbf{HealthBench}\\\textbf{Hard}}
& \shortstack{\textbf{Med-}\\\textbf{MCQA}}
& \shortstack{\textbf{PubMed-}\\\textbf{QA}}
& \shortstack{\textbf{MMLU-}\\\textbf{Med}}
& \textbf{GPQA} & \textbf{IFEval}
& \textbf{AIME24} & \textbf{AIME25} & \textbf{Mean} \\
\midrule
Qwen3-4B teacher & 6.38 & 57.26 & 76.60 & 78.32 & 42.42 & 86.88 & 54.00 & 43.33 & 55.65 \\
Medical Teacher & 39.06 & 59.07 & 70.00 & 80.09 & 57.68 & 73.38 & 47.33 & 42.00 & 58.58 \\
\midrule
M-OPD & 38.73 & 59.77 & 71.67 & 79.54 & 47.81 & 77.94 & \textbf{53.78} & 43.78 & 59.13 \\
GP-MOPD & \textbf{38.79} & \textbf{59.80} & 73.00 & \textbf{79.85} & 48.28 & 78.93 & 52.89 & 40.44 & 59.00 \\
\midrule
\rowcolor{uprow}\textbf{UP-MOPD} & 38.57 & 59.37 & \textbf{74.33} & 79.54 & \textbf{48.38} & \textbf{80.90} & 53.33 & \textbf{45.78} & \textbf{60.03} \\
\bottomrule
\end{tabular*}
\end{table}

\paragraph{Integration of Mathematics, Code, and Instruction Following.}
Table~\ref{tab:public} shows that both GP-MOPD and UP-MOPD improve the overall average over Open-MOPD. UP-MOPD achieves the highest average of 32.67, exceeding Open-MOPD and GP-MOPD by 1.82 and 1.04 points, respectively. It also achieves the best LiveCodeBench v5 score and code average, and matches the instruction-following teacher's IFEval score of 76.28.

\begin{table}[t]
\centering
\caption{Results on the public three-domain benchmark under the Open-MOPD evaluation protocol. Domain averages cover two tasks, and Total averages all six tasks. $^*$ marks results reported by \citet{ref3}; unmarked rows are our reproductions or proposed methods. RouteOPD uses three independently distilled students. Bold indicates the best result in each column.}
\label{tab:public}
\renewcommand{\arraystretch}{1.12}
\setlength{\tabcolsep}{3pt}
\footnotesize
\begin{tabularx}{\linewidth}{@{}>{\raggedright\arraybackslash}X*{7}{c}c*{2}{r}@{}}
\toprule
& \multicolumn{3}{c}{\textbf{Math}}
& \multicolumn{3}{c}{\textbf{Code}}
& \multicolumn{3}{c}{\textbf{Instruction Following}}
& \textbf{Overall} \\
\cmidrule(lr){2-4}\cmidrule(lr){5-7}\cmidrule(lr){8-10}\cmidrule(lr){11-11}
\textbf{Method} & \textbf{AIME24} & \textbf{AIME25} & \textbf{Avg.}
& \textbf{LCB5} & \textbf{LCB6} & \textbf{Avg.}
& \textbf{IFEval} & \textbf{IFBench} & \textbf{Avg.} & \textbf{Total} \\
\midrule
MixSFT & 13.33 & 17.78 & 15.56 & 15.57 & 17.71 & 16.64 & 69.44 & 16.00 & 42.72 & 24.97 \\
RFT$^*$ & 22.97 & 23.91 & 23.44 & 18.98 & 19.43 & 19.21 & 55.08 & 18.67 & 36.87 & 26.51 \\
MixRL$^*$ & 21.15 & 22.14 & 21.64 & 16.59 & 20.97 & 18.78 & 70.24 & 22.67 & 46.45 & 28.96 \\
ParamMerge-Avg$^*$ & 18.91 & 20.99 & 19.95 & 18.38 & 21.20 & 19.79 & 70.06 & 17.67 & 43.86 & 27.87 \\
ParamMerge-TA$^*$ & 21.93 & 22.76 & 22.34 & 21.74 & 23.14 & 22.44 & 71.53 & 21.67 & 46.60 & 30.46 \\
\midrule
RL-Math & \textbf{28.89} & 25.56 & \textbf{27.23} & 17.96 & 20.57 & 19.27 & 69.81 & 17.00 & 43.41 & 29.97 \\
RL-Code & 15.56 & 17.78 & 16.67 & 19.76 & \textbf{24.00} & 21.88 & 65.93 & 18.33 & 42.13 & 26.89 \\
RL-IF & 11.11 & 20.00 & 15.56 & 14.97 & 17.14 & 16.06 & \textbf{76.28} & \textbf{26.00} & \textbf{51.14} & 27.58 \\
\midrule
Naive M-OPD$^*$ & 20.92 & 21.60 & 21.26 & 17.53 & 20.99 & 19.26 & 68.61 & 18.67 & 43.64 & 28.05 \\
RouteOPD$^*$ & 22.34 & 23.96 & 23.15 & 22.28 & 21.14 & 21.71 & 75.60 & 24.00 & 49.80 & 31.55 \\
Open-MOPD & 21.11 & 23.33 & 22.22 & 22.16 & 20.00 & 21.08 & 74.49 & 24.00 & 49.25 & 30.85 \\
\midrule
GP-MOPD & 22.22 & \textbf{26.67} & 24.45 & 22.16 & 21.71 & 21.94 & 75.35 & 21.67 & 48.51 & 31.63 \\
\rowcolor{uprow}\textbf{UP-MOPD} & 26.67 & 24.44 & 25.56 & \textbf{25.15} & 21.14 & \textbf{23.15} & \textbf{76.28} & 22.33 & 49.31 & \textbf{32.67} \\
\bottomrule
\end{tabularx}
\end{table}

\subsection{Ablation and Analysis}
\label{sec:exp_mechanism}

We examine projection location and conflict handling in the medical--general setting (Table~\ref{tab:ablation}). M-OPD serves as the baseline without gradient clipping or projection, while M-OPD + Grad. Clip adds global gradient clipping. GP-MOPD projects the mixed gradient before AdamW. Update Rejection skips parameter updates that violate the same domain constraints used by UP-MOPD while retaining the optimizer-state update. Domain reweighting adjusts the mixture weights, with variants that either skip unresolved updates or select the candidate with the smallest remaining conflict.

\begin{table}[t]
\centering
\caption{Effect of intervention strategies in the medical--general setting, using the same evaluation protocol as Table~\ref{tab:1}. Mean averages the eight metrics. Bold indicates the best result in each column.}
\label{tab:ablation}
\renewcommand{\arraystretch}{1.08}
\scriptsize
\setlength{\tabcolsep}{3pt}
\begin{tabular*}{\linewidth}{@{\extracolsep{\fill}}lccccccccc@{}}
\toprule
& \multicolumn{4}{c}{\textbf{Medical}}
& \multicolumn{2}{c}{\textbf{General}}
& \multicolumn{2}{c}{\textbf{Math}} & \\
\cmidrule(lr){2-5}\cmidrule(lr){6-7}\cmidrule(lr){8-9}
\textbf{Method}
& \shortstack{\textbf{HealthBench}\\\textbf{Hard}}
& \shortstack{\textbf{Med-}\\\textbf{MCQA}}
& \shortstack{\textbf{PubMed-}\\\textbf{QA}}
& \shortstack{\textbf{MMLU-}\\\textbf{Med}}
& \textbf{GPQA} & \textbf{IFEval}
& \textbf{AIME24} & \textbf{AIME25} & \textbf{Mean} \\
\midrule
M-OPD & 38.73 & 59.77 & 71.67 & 79.54 & 47.81 & 77.94 & 53.78 & 43.78 & 59.13 \\
M-OPD + Grad. Clip & 38.31 & 59.42 & 73.47 & \textbf{80.04} & 48.35 & 73.20 & 53.78 & \textbf{45.78} & 59.04 \\
Reweighting w/ rejection & \textbf{39.11} & \textbf{60.00} & 73.27 & 79.94 & 45.25 & 76.77 & 52.89 & 44.89 & 59.02 \\
Reweighting w/o rejection & 38.16 & 59.49 & 72.87 & 79.77 & 47.44 & 78.99 & \textbf{54.67} & 45.11 & 59.56 \\
Update Rejection & 38.68 & 58.07 & 73.87 & 79.11 & 45.99 & 80.77 & 52.00 & 44.67 & 59.15 \\
GP-MOPD & 38.79 & 59.80 & 73.00 & 79.85 & 48.28 & 78.93 & 52.89 & 40.44 & 59.00 \\
\midrule
\rowcolor{uprow}\textbf{UP-MOPD} & 38.57 & 59.37 & \textbf{74.33} & 79.54 & \textbf{48.38} & \textbf{80.90} & 53.33 & \textbf{45.78} & \textbf{60.03} \\
\bottomrule
\end{tabular*}
\end{table}

\paragraph{Projection location and conflict handling.}
UP-MOPD achieves an eight-metric average of 60.03, compared with 59.00 for GP-MOPD and 59.15 for Update Rejection. It improves five of the eight metrics over GP-MOPD and seven over Update Rejection, and also achieves a higher average than both domain-reweighting variants. These comparisons support correcting the optimizer update as an effective way to integrate domain capabilities.

\begin{wrapfigure}[23]{l}{0.67\linewidth}
  \centering
  \vspace{-2pt} %
  \makebox[\linewidth][c]{%
    \hspace*{-20pt}%
    \includegraphics[width=\linewidth]{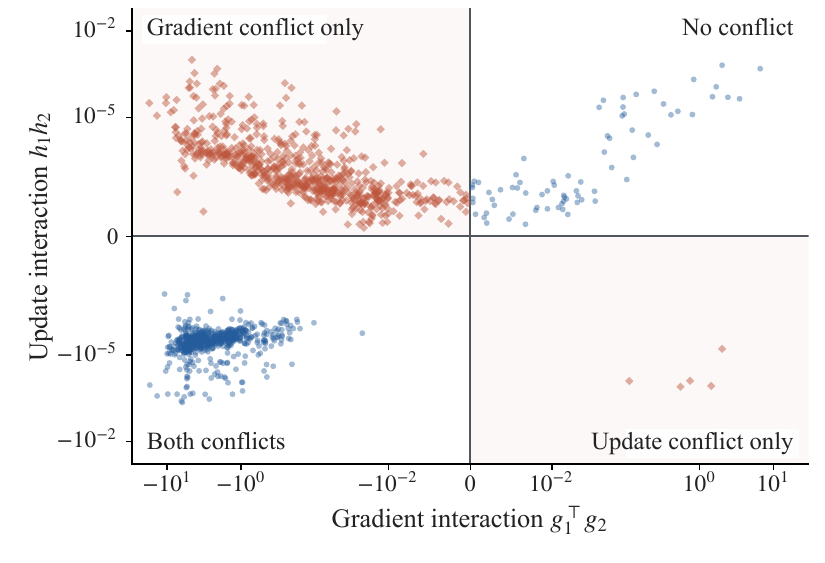}%
  }
  \caption{Gradient alignment and update conflict. Each point is a training step, with $g_1^\top g_2$ on the horizontal axis and $h_1h_2$ on the vertical axis. A negative $h_1h_2$ means the candidate update is predicted to lower one domain's loss and raise the other's. Gradient alignment alone does not determine this outcome.}
  \label{fig:grad-update}
  \vspace{-20pt} %
\end{wrapfigure}

\paragraph{Gradient and update checks can disagree.}
We analyze the 1,391 steps containing both domains in one epoch of the main training run. First-order constraint checks based on the mixed-gradient descent direction $-g_0$ and AdamW's candidate update disagree on 9.99\% of these steps. In 1.65\% of steps, the gradient check passes but the update check fails; 8.34\% show the opposite pattern. Figure~\ref{fig:grad-update} complements this analysis by visualizing the relationship between domain-gradient alignment and the first-order effects of optimizer updates on the two domain losses.

\begin{figure}[t]
    \centering
    \includegraphics[width=\linewidth]{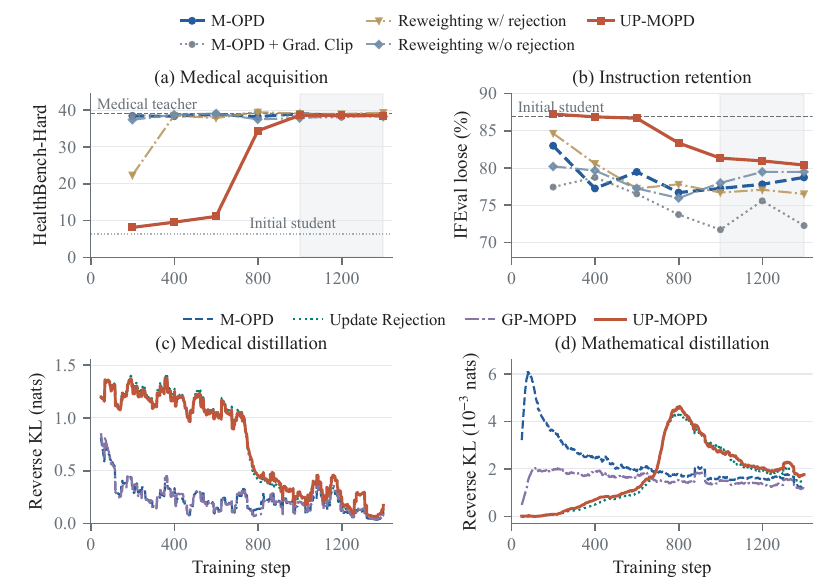}
    \caption{Capability and distillation dynamics. (a,b) HealthBench-Hard and
    IFEval-loose checkpoint scores; lines denote teacher or initial-student
    references and shading the late-training window. (c,d) Trailing 50-step
    means of domain-conditioned sampled reverse KL on each method's responses;
    steps lacking a domain and incomplete initial windows are omitted.}
    \label{fig:medical_dynamics}
\end{figure}

\paragraph{Training dynamics and capability retention.}
Figure~\ref{fig:medical_dynamics}(a,b) shows that UP-MOPD retains
instruction-following accuracy close to the initial student early in
training and later reaches medical performance comparable to M-OPD.
Its late-stage HealthBench-Hard score averages 38.57 versus 38.73 for
M-OPD, while its IFEval-loose accuracy is 80.90 versus 77.94 for M-OPD.
Panels (c,d) connect this capability development to the four methods'
sampled reverse-KL trajectories on medical and mathematical prompts.
Divergence from the general teacher on mathematical prompts is the
base-anchor loss. M-OPD and GP-MOPD reduce medical divergence early,
whereas UP-MOPD and Update Rejection maintain a lower base-anchor loss
for longer before substantially reducing medical divergence.
UP-MOPD also improves the eight-benchmark mean over Update Rejection
by 0.88 points (Table~\ref{tab:ablation}).

\section{Related Work}

\subsection{Knowledge Distillation and On-Policy Distillation}

Classical knowledge distillation transfers soft predictive distributions from
teachers or ensembles to a student \citep{ref1}. Generalized Knowledge
Distillation extends this supervision to student-generated sequences,
establishing on-policy distillation for language models \citep{ref2}.
Subsequent work studies self-distillation, reward design and extrapolation,
teacher entropy, training mechanisms, update geometry, and token teachability
\citep{ref16,ref17,ref18,ref19,ref20,ref21,ref22,xie2026step}. These methods
mainly alter supervision or trajectory construction; UP-MOPD instead constrains
the parameter displacement produced by a fixed distillation signal and shared
stateful optimizer.

\subsection{Multi-Teacher Capability Integration}

MiMo-V2-Flash and MOPD consolidate independently trained domain RL teachers
through on-policy distillation \citep{ref31,ref32}; GLM-5 uses cross-stage
distillation to recover capabilities from earlier SFT and RL checkpoints
\citep{ref33}. Open-MOPD addresses imbalance across mathematics, code, and
instruction following through effective optimization budgets \citep{ref3},
while related work uses privileged soft prompts \citep{ref23}, explores
evidence-aligned representation learning \citep{chen2026learning}, or studies
drift when top-$k$ supervision omits decision-critical tokens \citep{ref24}.
These approaches modify teacher organization, token supervision, or domain
allocation. UP-MOPD instead constrains the shared optimizer displacement after
gradient aggregation.

\subsection{Constrained Updates in Multi-Task and Continual Learning}

Multi-task optimization constructs shared directions through Pareto descent
(MGDA) \citep{ref25}, pairwise gradient projection (PCGrad) \citep{ref26},
cosine alignment (GradVac) \citep{ref34}, worst-task improvement (CAGrad)
\citep{ref27}, or adaptive norm balancing (GradNorm) \citep{ref28}. These
methods modify the gradient presented to the optimizer. UP-MOPD instead
constrains the candidate displacement produced by AdamW, with GP-MOPD providing
a matched gradient-space comparison in the same M-OPD setting.

Continual-learning methods likewise protect prior capabilities by constraining
new updates: GEM uses stored examples, while A-GEM uses an averaged reference
gradient \citep{ref29,ref30}. UP-MOPD instead constructs instantaneous
half-spaces from the active-domain gradients in the current logical batch and
requires no additional replay buffer. Its constraints are therefore local
first-order conditions, not guarantees against long-term forgetting.

\section{Conclusion}

We introduced UP-MOPD to correct a mismatch between gradient-space conflict
and the parameter displacement produced by adaptive optimizers in M-OPD. It
retains the current-step optimizer state induced by the original mixed
gradient, leaves feasible candidates unchanged, and applies the unique minimum
Euclidean correction satisfying all active-domain first-order constraints. A
domain-dimensional dual makes the projection practical at model scale. In
medical--general consolidation, UP-MOPD improves aggregate performance over
M-OPD, GP-MOPD, and update rejection while maintaining HealthBench-Hard; it
also achieves the highest six-task mean on the public
mathematics--code--instruction benchmark. Together, these results support
optimizer-induced displacement as a distinct intervention target for
multi-teacher capability integration.

\section*{AI Use Statement}

Generative AI tools were used to assist with manuscript drafting,
Chinese--English translation and language editing, secondary checking of
mathematical derivations, and the preparation and refinement of experimental
figures. All AI-assisted text, equations, citations, and visual materials were
reviewed and, where necessary, revised by the authors against the underlying
derivations, implementation, experiment logs, and source literature. The
authors take full responsibility for the research decisions, claims, and final
content of this paper.

\section*{Ethics Statement}

This work uses publicly released models and benchmark datasets under their
applicable licenses. No new human-subject data or private personally
identifiable information was collected. The medical experiments evaluate model
capabilities on research benchmarks and do not constitute clinical validation
or support deployment in healthcare. As with other capability-consolidation
methods, UP-MOPD may retain undesirable behaviors inherited from its teachers
or training data. Any deployment therefore requires task-specific safety
evaluation, license compliance, and appropriate human oversight.

\section*{Reproducibility Statement}

Section~2 specifies the M-OPD objective, GP-MOPD, and UP-MOPD, with formal
proofs in Appendix~\ref{app:theoretical_proofs}. Algorithm~\ref{alg:up_mopd}, domain-gradient
reconstruction, the dual solver, finite-precision commit verification, and
complexity accounting are provided in Appendix~\ref{app:implementation}.
Appendices~\ref{app:exp_training}--\ref{app:runtime} document the datasets,
released model and checkpoint identifiers, training mixtures, hyperparameters,
evaluation protocols, hardware, and runtime. The main tables and Appendix~\ref{app:experiments}
also report checkpoint aggregation rules and complete training trajectories.
Together, these details specify the method and experimental protocol required
to reproduce the reported results.

\clearpage
\appendix
\section{Limitations}

Our experiments cover capability integration with two and three
teachers. As the number of teachers grows, the cost of computing
separate domain gradients and the interactions among domain
constraints require further study. UP-MOPD also imposes local
first-order constraints at each training step; how these constraints
shape long-term capability retention remains an open question.

\section{Theoretical Proofs}
\label{app:theoretical_proofs}

\subsection{Proof of Proposition~\ref{prop:precondition}}
\label{app:proof_precondition}

Consider two domain gradients mixed with equal weights:

\[
g_1=(2,-1)^\top,\qquad g_2=(0,3)^\top,
\qquad g_0=\tfrac12(g_1+g_2)=(1,1)^\top.
\]

The unpreconditioned displacement $\Delta_{\mathrm{grad}}=-g_0=(-1,-1)^\top$ is a strict descent direction for both domains:

\[
g_1^\top\Delta_{\mathrm{grad}}=-1<0,
\qquad
g_2^\top\Delta_{\mathrm{grad}}=-3<0.
\]

Now take the positive diagonal preconditioner $D=\operatorname{diag}(1/10,1)$. Then

\[
\Delta_{\mathrm{pre}}=-Dg_0=(-1/10,-1)^\top,
\qquad
g_1^\top\Delta_{\mathrm{pre}}=4/5>0.
\]

Thus, positive diagonal preconditioning can turn a common descent direction into a displacement that increases one domain's loss to first order. The counterexample isolates the effect of coordinatewise scaling in optimizers such as Adam and AdamW. 

\subsection{Proof of Proposition~\ref{prop:projection} and Corollary~\ref{cor:mixed_first_order}}
\label{app:proof_projection}

The set $\mathcal C_t$ is the intersection of finitely many closed half-spaces and contains the zero vector; hence it is a nonempty closed convex cone. The squared-distance objective is continuous and coercive, so it attains a minimum on $\mathcal C_t$; its strong convexity makes that minimizer unique. This minimizer is the Euclidean projection of $\Delta\theta_0$ onto $\mathcal C_t$, and therefore satisfies the minimum-distance relation in Proposition~\ref{prop:projection} for every $\Delta\theta\in\mathcal C_t$. If $\Delta\theta_0\in\mathcal C_t$, the zero correction uniquely attains this lower bound.

Because $\mathcal C_t$ is polyhedral, the projection optimality condition $\Delta\theta_0-\Delta\theta^\star\in N_{\mathcal C_t}(\Delta\theta^\star)$ yields multipliers $\lambda^\star\ge0$ satisfying stationarity and complementary slackness.

When $K=G^\top G$ is singular, the dual coefficients may admit multiple optimal representations. The objective in Eq.~(\ref{eq:up_dual}) is equivalent to $\tfrac12\|G\lambda-\Delta\theta_0\|_2^2-\tfrac12\|\Delta\theta_0\|_2^2$. Let $\lambda_1$ and $\lambda_2$ be dual optima. If $G\lambda_1\ne G\lambda_2$, their feasible midpoint would achieve a strictly smaller objective by strict convexity of the squared norm, contradicting optimality. Therefore $G\lambda_1=G\lambda_2$: singularity affects only the representation of the dual coefficients, not the correction or the final projected displacement.

By Eq.~(\ref{eq:up_correction}) and complementary slackness in Eq.~(\ref{eq:up_kkt}),

\[
\Delta\theta_0-\Delta\theta^\star=G\lambda^\star,
\qquad
(\Delta\theta^\star)^\top(\Delta\theta_0-\Delta\theta^\star)
=(\lambda^\star)^\top G^\top\Delta\theta^\star=0.
\]

Expanding $\|\Delta\theta_0\|_2^2=\|\Delta\theta^\star+(\Delta\theta_0-\Delta\theta^\star)\|_2^2$ gives the Pythagorean identity. A positive rescaling of any constraint normal leaves the half-space $g_d^\top\Delta\theta\le0$ unchanged and therefore preserves both the feasible cone and its unique projection. Because $q_d>0$, domain-wise feasibility also gives

\[
g_0^\top\Delta\theta^\star
=\sum_dq_d g_d^\top\Delta\theta^\star\le0.
\]

Finally, if $L_d$ has a $\beta_d$-Lipschitz gradient, the descent lemma and $g_d^\top\Delta\theta^\star\le0$ imply

\[
L_d(\theta_t+\Delta\theta^\star)
\le L_d(\theta_t)+g_d^\top\Delta\theta^\star
+\frac{\beta_d}{2}\|\Delta\theta^\star\|_2^2
\le L_d(\theta_t)+\frac{\beta_d}{2}\|\Delta\theta^\star\|_2^2.
\]

The bound leaves a second-order remainder proportional to the squared displacement norm. At a tight first-order constraint, this remainder allows a positive loss change for a finite step. 

\section{Implementation and Complexity Details}
\label{app:implementation}

Algorithm~\ref{alg:up_mopd} summarizes the complete update. The
\textsc{MaterializeAndVerify} subroutine is specified in
Appendix~\ref{app:finite_precision}.

\begin{algorithm}[H]
\caption{One training step of UP-MOPD}
\label{alg:up_mopd}
\small
\begin{algorithmic}[1]
\Require Parameters $\theta_t$, optimizer state $\mathcal M_{t-1}$, sampled batch
\State Compute the mixed gradient $g_0$
\If{$D_t<2$}
    \State \Return $\operatorname{Opt}_t(\theta_t,\mathcal M_{t-1};g_0)$
\EndIf
\State Replay the same batch at $\theta_t$ to obtain $G=[g_d]_{d\in\mathcal A_t}$
\State Save the pre-update master parameters $\bar\theta_t\gets\theta_t$
\State $(\theta_{t+1}^{0},\mathcal M_t)\gets
       \operatorname{Opt}_t(\theta_t,\mathcal M_{t-1};g_0)$
\State $\Delta\theta_0\gets\theta_{t+1}^{0}-\bar\theta_t$,
       $h\gets G^\top\Delta\theta_0$
\State $\Delta\theta^\star\gets\Delta\theta_0$
\If{$h_d>0$ for any $d\in\mathcal A_t$}
    \State Compute $K\gets G^\top G$ and solve Eq.~(\ref{eq:up_dual})
    \State $\Delta\theta^\star\gets\Delta\theta_0-G\lambda^\star$
\EndIf
\State $\widehat{\Delta\theta}\gets
       \Call{MaterializeAndVerify}{\bar\theta_t,\Delta\theta^\star,G}$
\State \Return $(\bar\theta_t+\widehat{\Delta\theta},\mathcal M_t)$
\end{algorithmic}
\end{algorithm}

\subsection{Generation Statistics and Update Behavior}
\label{app:four_method_dynamics}

Table~\ref{tab:four_method_outcomes} supplements the training dynamics
in Section~\ref{sec:exp_mechanism} with generation statistics.
Across the four methods, mean response lengths range from 4,056 to
4,089 tokens and truncation rates from 9.20\% to 9.63\% over the final
200 plotted training steps.

\begin{table}[H]
    \centering
    \caption{Capability scores and generation behavior.
    HB-Hard denotes HealthBench-Hard; IFEval uses prompt-level loose accuracy.
    Mean averages eight benchmarks. Tokens and Trunc. are mean response
    length and truncation rate over the last 200 plotted training steps.}
    \label{tab:four_method_outcomes}
    \small
    \setlength{\tabcolsep}{4pt}
    \begin{tabular*}{\linewidth}{@{\extracolsep{\fill}}lrrrrr@{}}
        \toprule
        Method & HB-Hard & IFEval & Mean & Tokens & Trunc. (\%) \\
        \midrule
        M-OPD & 38.73 & 77.94 & 59.13 & 4,067 & 9.20 \\
        Update Rejection & 38.68 & 80.77 & 59.15 & 4,071 & 9.48 \\
        GP-MOPD & \textbf{38.79} & 78.93 & 59.00 & 4,056 & 9.63 \\
        UP-MOPD & 38.57 & \textbf{80.90} & \textbf{60.03} & 4,089 & 9.20 \\
        \bottomrule
    \end{tabular*}
\end{table}

\paragraph{Update behavior.}
Update Rejection cancels 645 candidates, each predicting a medical
loss decrease but an increase in the base-anchor loss.

\subsection{Projecting Rejected Updates}
\label{app:rejection_counterfactual}

\begin{figure}[H]
    \centering
    \includegraphics[width=\linewidth]{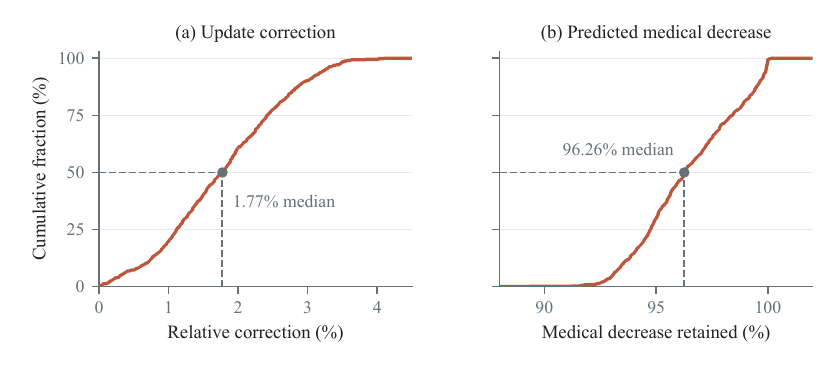}
    \caption{Offline projection of 645 rejected candidates.
    Cumulative distributions show relative correction size (left)
    and retained predicted medical loss decrease (right).
    Dashed lines mark medians.}
    \label{fig:appendix_rejection_counterfactual}
\end{figure}

We take all 645 candidates discarded by Update Rejection and compute
their UP-MOPD projections offline from the logged gradient statistics.
All yield nonzero projections satisfying both domains' first-order constraints.
The relative correction norm has a median of 1.77\% and a
95th percentile of 3.25\%.
The projected updates retain a median of 96.26\% of the predicted
medical loss decrease (Figure~\ref{fig:appendix_rejection_counterfactual}).
Projection thus removes the predicted increase in base-anchor loss
while retaining most of the predicted medical decrease.

\paragraph{Metric definitions.}
For candidate $\Delta\theta_0$ and projection $\Delta\theta^\star$,
we report the relative correction $r_{\mathrm{corr}}$ and medical
decrease retention $r_{\mathrm{med}}$:
\begin{equation}
r_{\mathrm{corr}}
=
\frac{\|\Delta\theta^\star-\Delta\theta_0\|}
     {\|\Delta\theta_0\|},
\qquad
r_{\mathrm{med}}
=
\frac{h_{\mathrm{med}}(\Delta\theta^\star)}
     {h_{\mathrm{med}}(\Delta\theta_0)}.
\end{equation}
Here $h_d(\Delta)=g_d^\top\Delta$ predicts the first-order change in
domain $d$'s loss.
Since $h_{\mathrm{med}}(\Delta\theta_0)<0$ for every candidate,
$r_{\mathrm{med}}$ measures the fraction of predicted medical
decrease retained.
Medical decrease retention ranges from 89.35\% to 100.14\%;
values above 100\% indicate a slightly larger predicted decrease
after projection.
\subsection{Domain-Gradient Replay and Consistency Check}
\label{app:domain_replay}

Equation~(\ref{eq:mopd_domain_loss}) gives the sampled-token estimator. In the public implementation, teacher and student log probabilities on a shared top-$k$ support define the corresponding dense token estimator; restricting either estimator to domain $d$ yields $L_d$ and $g_d$ used by the projection. The response weights $\omega_i$ unify the two normalization schemes. Setting $\omega_i=1/M_i$ first averages tokens within each response and then averages responses within a domain; setting $\omega_i=1$ yields a domain-level token mean. Standard backpropagation of the mixed objective produces $g_0$, while domain-wise replay over the same logical batch produces the individual domain gradients. The implementation computes

\[
e_{\mathrm{dec}}
=\frac{\|g_0-g_{\mathrm{rec}}\|_2}
{\max(\|g_0\|_2,10^{-30})},
\qquad
g_{\mathrm{rec}}=\sum_{d\in\mathcal A_t}q_dg_d,
\tag{C.1}
\]

and, when the check is enabled, requires $e_{\mathrm{dec}}\le\tau_{\mathrm{dec}}$. Domain replay and mixed-objective backpropagation use the same sampled data, stochastic state, and gradient processing. If global scalar clipping is applied, its factor is computed from the mixed gradient and then applied uniformly to every domain contribution; domains are not clipped independently.

\subsection{Dual Active Sets, Scale Normalization, and Distributed Computation}
\label{app:dual_solver}

For a candidate active set $A$, the nonzero dual variables satisfy

\[
K_{AA}\lambda_A=h_A,\qquad \lambda_{\bar A}=0.
\tag{C.2}
\]

When $K_{AA}$ is nonsingular, we solve this linear system directly. When the domain gradients are linearly dependent, least squares produces one representative dual solution, after which we check dual nonnegativity, primal feasibility, and the KKT residual. Because the number of domains in our setting is small, we enumerate all active sets; larger settings can instead use a general nonnegative quadratic-programming solver.

To reduce ill-conditioning caused by differences in domain-gradient scale, define

\[
s_d=\begin{cases}\|g_d\|_2,&\|g_d\|_2>0,\\1,&\|g_d\|_2=0,\end{cases}
\quad
\bar g_d=g_d/s_d,\quad
\mu_d=s_d\lambda_d,\quad
\bar K_{ij}=\bar g_i^\top\bar g_j,\quad
\bar h_d=h_d/s_d.
\tag{C.3}
\]

After obtaining $\mu_d^\star$, we recover the original dual variable as $\lambda_d^\star=\mu_d^\star/s_d$. Each parameter shard accumulates $K_{ij}$ and $h_i$ in a streaming manner, followed by an all-reduce over the resulting global scalars. All shards share the same $\lambda^\star$ and reconstruct $\Delta\theta^\star$ locally, without gathering full parameter vectors.

As an optional extension, Eq.~(\ref{eq:up_projection}) can be defined under a positive-definite weighted norm. In that case, stationarity becomes $\Delta\theta^\star=\Delta\theta_0-M^{-1}G\lambda^\star$, and the dual Gram matrix becomes $G^\top M^{-1}G$. The dual dimension remains unchanged, although the resulting projection depends on the metric. The method and all experiments in this paper use the Euclidean metric, $M=I$.

\subsection{Optimizer State and Finite-Precision Updates}
\label{app:finite_precision}

Each logical step invokes the original optimizer once. UP-MOPD replaces its candidate parameter update while retaining the state produced by the mixed gradient:

\[
\theta_{t+1}=\theta_t+\widehat{\Delta\theta},\qquad
\mathcal M_t=\operatorname{StateUpdate}(\mathcal M_{t-1},g_0^{\mathrm{opt}}).
\tag{C.4}
\]

The current optimizer state is updated using the shared, optionally clipped mixture $g_0^{\mathrm{opt}}$, and subsequent gradients are evaluated along the projected parameter trajectory. The state update is retained even when the fallback below sets the parameter displacement to zero.

Forward and backward passes use BF16, optimizer master parameters use FP32, and the projection target is formed in FP64. Let $Q_b$ denote rounding to storage type $b$. The stored parameters, realized displacement, and predicted domain loss changes are

\[
\theta_{t+1}^{\mathrm{mat}}=Q_b(\theta_t+\Delta\theta^\star),\qquad
\widehat{\Delta\theta}=\theta_{t+1}^{\mathrm{mat}}-\theta_t,\qquad
\widehat h_d=g_d^\top\widehat{\Delta\theta}.
\tag{C.5}
\]

The result is committed only if $\widehat h_d\le0$ for every domain. If rounding reintroduces a violation, correction round $r$ solves

\[
\begin{aligned}
v_r&=\max_{d:\|g_d\|_2>0}
\frac{(\widehat h_d^{(r)})_+}{\|g_d\|_2},\\
c_r^\star&=\arg\min_c\tfrac12\|c\|_2^2\\
&\text{s.t.}\quad
g_d^\top(\widehat{\Delta\theta}^{(r)}+c)
\le-\gamma_rv_r\|g_d\|_2,\qquad
\gamma_r\in\{2,4\}.
\end{aligned}
\tag{C.6}\label{eq:rounding_correction}
\]

Equation~(\ref{eq:rounding_correction}) uses the same constraint normals and dual solver as Eq.~(\ref{eq:up_projection}). The negative margin adapts to the previous round's rounding residual. We allow at most two correction rounds, using $\gamma_r=2$ and then $4$, and round and verify the parameters after each round.

If the correction problem is infeasible, produces a non-finite result, requires $\|c_r^\star\|_2>\|\widehat{\Delta\theta}^{(r)}\|_2$, or fails final verification, we restore the pre-update parameters and commit a zero displacement. A zero returned directly by Eq.~(\ref{eq:up_projection}) is accepted as the optimal projection.

The committed displacement satisfies the domain constraints at the verification precision. The minimum-distance property in Proposition~\ref{prop:projection} applies to the exact projection.

\subsection{Computational and Communication Complexity}
\label{app:complexity}

Let $D=|\mathcal A_t|$ and let $P$ be the number of trainable parameters. On a multi-domain step, domain replay requires $D$ gradient passes in addition to mixed-objective backpropagation; the single-domain fast path requires none. Streaming construction of the Gram matrix and candidate changes costs $O(D^2P)$, and each materialization or correction round costs $O(DP)$. Distributed communication consists of $O(D^2)$ scalars, while each parameter shard requires $O((D+1)P_{\mathrm{local}})$ additional storage. Active-set enumeration has worst-case complexity $O(2^DD^3)$, but its dimension is governed by the number of domains rather than the number of model parameters.

\section{Additional Experimental Details and Results}
\label{app:experiments}

\subsection{Experimental Design and Training Settings}
\label{app:exp_training}

\paragraph{Medical specialization with capability retention.}
The medical setting asks whether a student can acquire a specialist
teacher's capabilities while retaining those of its initialization.
We initialize the student from Qwen3-4B-Instruct-2507 and use the same
checkpoint as the frozen general teacher. This teacher anchors training
on mathematical prompts, and IFEval evaluates instruction-following
retention after medical distillation.
We combine 17,398 mathematical prompts with 5,166 prompts routed to the
medical teacher, shuffle the combined pool, and train for one epoch.

The medical teacher is based on Qwen3-4B and trained following the medical reinforcement learning recipe of \citet{zhu2026conrubmed}.
Each generated response is supervised by the teacher assigned to its
prompt's domain. We average token losses within each response and then
average over responses. Thus, $q_d=N_d/N$, where $N_d$ is the number of
responses assigned to domain $d$ in a logical batch of $N$ responses.
The sampled-token advantage is the assigned teacher's log probability
minus the student's log probability. No external scalar reward or
advantage normalization is applied.

\paragraph{Reweighting controls.}
Both reweighting baselines test mixtures $\alpha g_1+(1-\alpha)g_2$
over $\alpha\in\{0,1/16,\ldots,1\}$, including the original batch mixture.
Each candidate is evaluated through AdamW using the same pre-step optimizer
state. Among candidates satisfying both domain constraints, they select
the mixture closest to the original weight. If none is feasible,
Reweighting w/ rejection discards the update; Reweighting w/o rejection
selects the candidate with the smallest maximum positive domain change,
normalized by the corresponding domain-gradient norm.

\paragraph{Integration of three specialist teachers.}
The public setting asks whether one student can combine the mathematics,
code, and instruction-following capabilities of three specialist teachers.
It extends the comparison to three active domains, a different model
family, and a verl implementation.
We initialize from the released SmolLM3-3B MixSFT checkpoint and use
the frozen RL-Math, RL-Code, and RL-IF teachers from Open-MOPD~\citep{ref3}.
Table~\ref{tab:public_assets} specifies the released models and prompt pool.
All models share the repository prefix
\texttt{BytedTsinghua-SIA/Open-MOPD-SmolLM3-3B-}.
The student uses the \texttt{MixSFT} release, corresponding to the fourth
SFT epoch. The mixed training file is
\texttt{rl\_prompt\_mix/train.parquet} in
\texttt{BytedTsinghua-SIA/Open-MOPD-Data}.

\begin{table}[htbp]
\centering
\small
\caption{Released teachers and training data in the public setting.
Model suffixes follow the shared repository prefix given in the text.
Teacher steps identify the checkpoints described in their model cards.
The released prompt pool excludes code records identified as LiveCodeBench.}
\label{tab:public_assets}
\begin{tabularx}{\linewidth}{@{}llrXr@{}}
\toprule
Domain & Teacher suffix & Step & Prompt source & Prompts \\
\midrule
Mathematics & \texttt{RL-Math} & 100 & DAPO-Math-17k & 17,917 \\
Code & \texttt{RL-Code} & 180 & DeepCoder-Preview-Dataset & 23,667 \\
Instruction following & \texttt{RL-IF} & 2,440 & Nemotron-IF-RL-46k & 45,347 \\
\midrule
Total & & & & 86,931 \\
\bottomrule
\end{tabularx}
\end{table}

GP-MOPD and UP-MOPD share a domain-routed, dense distillation objective.
Teacher and student log probabilities are evaluated on the same
student-selected support of 256 tokens per position. The token-mean reducer uses the same
microbatch normalization in mixed and domain-replay passes, so the
domain contributions sum to the original mixed gradient.
Each run of GP-MOPD and UP-MOPD processes 679 batches of 128 prompts:
17,914 mathematics, 23,663 code, and 45,335 instruction-following prompts. Tables~\ref{tab:hyper_medical} and~\ref{tab:hyper_public} give the
training hyperparameters. Both settings use a constant learning rate
without warmup and no additional KL or entropy regularization.

\begin{table}[H]
\centering
\caption{Training hyperparameters for medical and general capability integration.}
\label{tab:hyper_medical}
\footnotesize
\renewcommand{\arraystretch}{1.15}
\setlength{\tabcolsep}{5pt}
\begin{tabularx}{\linewidth}{@{}>{\raggedright\arraybackslash}p{0.25\linewidth}>{\raggedright\arraybackslash}X@{}}
\toprule
\textbf{Hyperparameter} & \textbf{Value} \\
\midrule
Student & Qwen3-4B-Instruct-2507 \\
Teachers & Qwen3-4B-Instruct-2507 and Medical Teacher; both frozen, with supervision routed by prompt domain \\
Training framework & Megatron \\
Training data & 22,564 prompts: 17,398 mathematical prompts and 5,166 prompts assigned to the medical teacher; combined and randomly shuffled \\
Training duration & 1 epoch; 1,410 steps \\
Batch size & 16 prompts per step; 4 responses per prompt (64 samples) \\
Maximum response length & 8,192 tokens; no-thinking mode \\
Optimizer parameters & Learning rate $10^{-6}$; $(\beta_1,\beta_2)=(0.9,0.98)$; weight decay 0.1 \\
Global gradient clipping & Disabled (threshold 0) \\
Distillation signal & Sampled-token teacher--student log-ratio; no advantage normalization \\
Rollout & SGLang; temperature 1.0 \\
Parallelism & Training: 2 GPUs, tensor parallelism 2 with sequence parallelism; rollout: 4 GPUs \\
Projection threshold & $\epsilon=0$ \\
\bottomrule
\end{tabularx}
\end{table}

\begin{table}[H]
\centering
\caption{Training hyperparameters for the public mathematics, code, and instruction-following benchmark.}
\label{tab:hyper_public}
\footnotesize
\renewcommand{\arraystretch}{1.15}
\setlength{\tabcolsep}{5pt}
\begin{tabularx}{\linewidth}{@{}>{\raggedright\arraybackslash}p{0.25\linewidth}>{\raggedright\arraybackslash}X@{}}
\toprule
\textbf{Hyperparameter} & \textbf{Value} \\
\midrule
Student & SmolLM3-3B initialized from MixSFT \\
Teachers & Frozen RL-Math, RL-Code, and RL-IF releases (Table~\ref{tab:public_assets}) \\
Training framework & verl; public data and checkpoints from Open-MOPD~\citep{ref3} \\
Training data & 86,931 prompts mixed across the three domains \\
Training duration & 1 epoch; 679 full batches \\
Batch size & 128 prompts per step; mini-batch size 128; micro-batch size 1 per GPU; 8 GPUs \\
Maximum sequence lengths & Prompt: 2,048 tokens; response: 16,384 tokens (2,048 for instruction following) \\
Optimizer & AdamW; learning rate $10^{-6}$; $(\beta_1,\beta_2)=(0.9,0.999)$; weight decay 0.01 \\
Global gradient clipping & Disabled (threshold 0) \\
Distillation signal & Dense token-level distillation on student top-256 support; token mean within each microbatch \\
Rollout & vLLM; 1 response per prompt; temperature 1.0; top-$p$ 1.0 \\
Update projection & Hard projection (\texttt{adam\_project\_hard}); $\epsilon=0$ \\
Random seeds & Data and rollout: 1 \\
\bottomrule
\end{tabularx}
\end{table}

\subsection{Evaluation Protocols}
\label{app:evaluation}

\paragraph{Medical and general benchmarks.}
For the medical setting, we disable thinking and use a sampling
temperature of 0.6. Table~\ref{tab:medical_eval} lists the
generation limits and repetitions. MedMCQA uses five fixed demonstrations;
the other benchmarks use zero-shot prompts.
For GPQA and AIME, avg@5 is mean correctness across five completions per
question. IFEval uses prompt-level loose accuracy. MMLU-Med averages eight
subjects with equal weight: anatomy, clinical knowledge, college biology,
college medicine, medical genetics, nutrition, professional medicine,
and virology.
Student results average the three regularly saved late-training
checkpoints; teacher results evaluate the corresponding fixed checkpoint.

\begin{table}[htbp]
\centering
\small
\caption{Generation settings for the medical evaluation.
Limits count generated tokens. Repetitions are completions per question.}
\label{tab:medical_eval}
\begin{tabular*}{\linewidth}{@{\extracolsep{\fill}}lrrr@{}}
\toprule
Benchmark & Token limit & Repetitions & Demonstrations \\
\midrule
HealthBench-Hard & 16,384 & 1 & 0 \\
GPQA-Diamond, AIME24/25 & 8,192 & 5 & 0 \\
IFEval, MMLU-Med & 8,192 & 1 & 0 \\
MedMCQA & 32,768 & 1 & 5 \\
PubMedQA & 32,768 & 1 & 0 \\
\bottomrule
\end{tabular*}
\end{table}

\paragraph{Public three-domain benchmarks.}
All models in Table~\ref{tab:public} follow the Open-MOPD evaluation
protocol. Generation uses vLLM with temperature 0.6, top-$p$ 0.95,
top-$k$ 20, a 30,000-token generation limit, and a 32,768-token context
limit. We use the released tokenizer's chat template without overriding
thinking mode. AIME24/25 use mean@64, LiveCodeBench v5/v6 use mean@10,
and IFEval and IFBench use mean@1. The evaluation sets contain 30, 30,
541, 167, 175, and 300 prompts, respectively. AIME uses final-answer
correctness, LiveCodeBench uses test-case execution, and IFEval and
IFBench use strict prompt-level accuracy. Each domain mean averages its
two benchmarks, and the overall mean averages all six. Rows quoted from
\citet{ref3} and our reproduced models therefore use the same benchmark
protocol.

\subsection{Computational Resources and Runtime}
\label{app:runtime}

UP-MOPD's mean logged step time is 1.42 times that of M-OPD;
GP-MOPD and Update Rejection have similar runtimes to UP-MOPD
(Table~\ref{tab:runtime}).
Logged step time sums training time and the interval between training
calls, which includes waiting for rollouts.
Actor training time covers forward and backward computation,
domain-gradient processing, and optimizer updates.

\begin{table}[htbp]
\centering
\caption{Runtime in the medical--general setting on eight NVIDIA H200 GPUs.
Values are means $\pm$ standard deviations in seconds per step, using the
same 200-step late-training window for all four methods.
Actor training is included in the logged step time.}
\label{tab:runtime}
\small
\renewcommand{\arraystretch}{1.10}
\begin{tabular*}{\linewidth}{@{\extracolsep{\fill}}lrr@{}}
\toprule
\textbf{Method} & \textbf{Logged step time (s)} & \textbf{Actor training (s)} \\
\midrule
M-OPD & $66.55 \pm 6.58$ & $11.57 \pm 1.65$ \\
Update Rejection & $92.58 \pm 9.45$ & $37.49 \pm 5.71$ \\
GP-MOPD & $92.94 \pm 9.69$ & $38.00 \pm 5.77$ \\
UP-MOPD & $94.76 \pm 10.08$ & $39.69 \pm 5.82$ \\

\bottomrule
\end{tabular*}
\end{table}

\Needspace{12\baselineskip}
\subsection{Public-Benchmark Training Trajectory}
Mathematics reaches its highest recorded score earlier in training,
whereas code peaks at the final checkpoint.
The final checkpoint also has the highest six-task mean
(Table~\ref{tab:trajectory}).

\begin{table}[htbp]
\centering
\caption{UP-MOPD training trajectory in the public three-domain setting.}
\label{tab:trajectory}
\renewcommand{\arraystretch}{1.10}
\footnotesize
\begin{tabular*}{\linewidth}{@{\extracolsep{\fill}}lrrrr@{}}
\toprule
\textbf{Step} & \textbf{Math} & \textbf{Code} & \textbf{IF} & \textbf{Total} \\
\midrule
100 & 18.89 & 21.24 & \textbf{50.58} & 30.24 \\
200 & \textbf{26.67} & 20.44 & 49.48 & 32.20 \\
300 & 23.33 & 19.87 & 48.53 & 30.57 \\
400 & 19.45 & 21.33 & 48.49 & 29.75 \\
500 & 23.89 & 21.93 & 47.54 & 31.12 \\
600 & 22.22 & 21.94 & 49.12 & 31.09 \\
\textbf{679} & 25.56 & \textbf{23.15} & 49.31 & \textbf{32.67} \\
\bottomrule
\end{tabular*}
\end{table}

\begin{figure}[t]
\centering
\includegraphics[width=\linewidth]{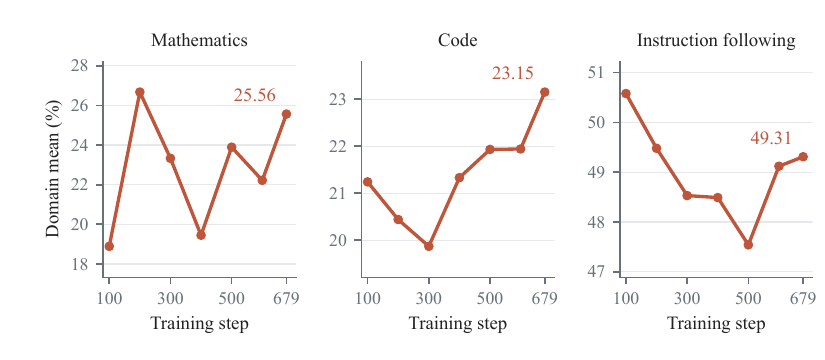}
\caption{UP-MOPD training trajectory in the public three-domain setting. Each panel shows a domain mean over two benchmarks. The vertical scales differ across panels.}
\label{fig:trajectory}
\end{figure}

\subsection{Sample-Level Paired Analysis in the Medical--General Setting}
\label{app:paired_analysis}

\paragraph{Instruction-following retention.}
The 2.96-point IFEval gain mainly comes from retaining the initial
student's instruction-following ability (Table~\ref{tab:ifeval_retention}).
The initial student passes 470 of 541 prompts under loose scoring.
Across the three late-training checkpoints, M-OPD fails an average
of 69.3 of these prompts, compared with 50.3 for UP-MOPD.
Newly passed prompts average 21.0 for M-OPD and 18.0 for UP-MOPD.
UP-MOPD retains more of the initial passes at all three checkpoints.

\begin{table}[H]
\centering
\small
\caption{IFEval retention across three late-training checkpoints.
The initial student passes 470 of 541 prompts under loose scoring.
Retained and lost prompts partition these 470 initial passes; newly
passed prompts come from the 71 initial failures. Counts and accuracies
are means over the three checkpoints.}
\label{tab:ifeval_retention}
\begin{tabular*}{\linewidth}{@{\extracolsep{\fill}}lrrrr@{}}
\toprule
Method & Retained & Lost & Newly passed & Accuracy (\%) \\
\midrule
M-OPD & 400.7 & 69.3 & 21.0 & 77.94 \\
UP-MOPD & 419.7 & 50.3 & 18.0 & 80.90 \\
\bottomrule
\end{tabular*}
\end{table}

\paragraph{Paired benchmark comparison.}
At the middle late-training checkpoint, UP-MOPD passes 38 IFEval prompts
that M-OPD fails and fails 21 that M-OPD passes.
Table~\ref{tab:paired} reports the paired comparison across all eight
benchmarks. None of the differences is significant after
Benjamini--Hochberg correction~\citep{ref15}.

\begin{table}[H]
\centering
\caption{Paired comparison at the middle late-training checkpoint.
IFEval uses prompt-level loose accuracy. Differences are UP-MOPD minus
M-OPD, in score points. Intervals are pointwise 95\% paired bootstrap
confidence intervals; $p$-values are corrected across all eight metrics.}
\label{tab:paired}
\renewcommand{\arraystretch}{1.10}
\footnotesize
\begin{tabular*}{\linewidth}{@{\extracolsep{\fill}}lrrrrr@{}}
\toprule
\textbf{Metric} & \textbf{M-OPD} & \textbf{UP-MOPD}
& \textbf{Difference} & \textbf{95\% CI}
& \shortstack{\textbf{BH-adjusted}\\$p$} \\
\midrule
HealthBench-Hard & 38.51 & 38.63 & $+0.12$ & $[-1.34,1.60]$ & 1.0000 \\
GPQA & 49.80 & 48.18 & $-1.62$ & $[-4.65,1.31]$ & 0.5279 \\
IFEval & 77.82 & 80.96 & $+3.14$ & $[0.37,5.91]$ & 0.1454 \\
MedMCQA & 59.67 & 58.71 & $-0.96$ & $[-1.77,-0.12]$ & 0.1454 \\
PubMedQA & 72.00 & 74.40 & $+2.40$ & $[-0.20,5.00]$ & 0.2564 \\
MMLU-Med & 79.41 & 79.44 & $+0.03$ & $[-1.34,1.44]$ & 1.0000 \\
AIME24 & 54.67 & 54.00 & $-0.67$ & $[-7.33,5.33]$ & 1.0000 \\
AIME25 & 42.67 & 47.33 & $+4.67$ & $[-0.67,11.33]$ & 0.4054 \\

\bottomrule
\end{tabular*}
\end{table}

We use exact McNemar tests for IFEval, MedMCQA, and PubMedQA, and paired
sign-flip tests for HealthBench-Hard, GPQA, and AIME.
MMLU-Med uses a subject-stratified sign-flip test and bootstrap,
with equal subject weights. All tests are two-sided.
Bootstrap confidence intervals and permutation tests use 4,000 repetitions.
GPQA and AIME average the five generations for each question before
pairing. Benjamini--Hochberg correction covers all eight metrics.

On HealthBench-Hard, 366 of the 1,000 paired scores increase, 288 remain
unchanged, and 346 decrease. The correlation between response-length
change and score change is $-0.049$, indicating a weak linear association
between the two changes.

\begin{figure}[H]
  \centering
  \includegraphics[width=\textwidth]{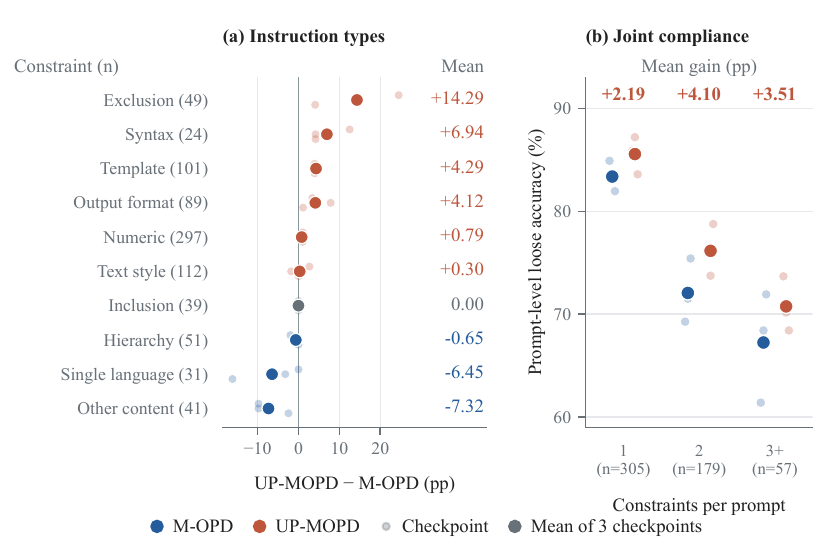}
  \caption{\textbf{Instruction-following behavior in the medical setting.}
  (a) UP-MOPD minus M-OPD instruction-level loose accuracy for all ten evaluator-defined constraint categories; parentheses give instruction counts.
  (b) Prompt-level loose accuracy grouped by the number of constraints in a prompt; labels above the groups give the mean gain in percentage points.
  Faint points show three matched late-training checkpoints from one run per method; solid points show their arithmetic mean.}
  \label{fig:ifeval-constraints}
\end{figure}

\Needspace{8\baselineskip}
\paragraph{Instruction type and constraint count.}
Figure~\ref{fig:ifeval-constraints} shows the largest mean gains for
exclusion constraints (+14.29 points), syntax (+6.94), templates (+4.29),
and output formats (+4.12). Other-content and single-language constraints
decrease by 7.32 and 6.45 points, respectively.
UP-MOPD improves mean prompt-level accuracy in all three groups defined
by the number of constraints, with larger gains when prompts contain
multiple constraints. Table~\ref{tab:ifeval-cases} shows two prompts where
UP-MOPD preserves compliance with the requested bullet count and word
limit.

\begin{table}[H]
\centering
\small
\setlength{\tabcolsep}{4pt}
\begin{tabularx}{\textwidth}{@{}p{.25\textwidth}XX@{}}
\toprule
Constraint & M-OPD & UP-MOPD \\
\midrule
Exactly three Markdown bullet points
& Repeats the three-item list after ``Final concise version with desert emphasis:''. \textbf{Loose: fail.}
& Produces one three-item list. \textbf{Loose: pass.} \\
\addlinespace[5pt]
At most 100 words and at least two highlighted sections
& 115 words and two highlighted spans. Highlighting passes, but the length constraint fails. \textbf{Loose: fail.}
& 99 words and three highlighted spans. Both constraints pass. \textbf{Loose: pass.} \\
\bottomrule
\end{tabularx}
\caption{\textbf{Examples of constraint compliance.} The initial student passes both prompts; distilled outputs are from the middle late-training checkpoint. Counts are computed from complete responses; word counts follow IFEval's regular-expression tokenizer.}
\label{tab:ifeval-cases}
\end{table}

\end{document}